# InstantFamily: Masked Attention for Zero-shot Multi-ID Image Generation

Chanran Kim
SK Telecom
Seoul, Republic of Korea
chanrankim@sk.com

Jeongin Lee
SK Telecom
Seoul, Republic of Korea
jeonginlee@sk.com

Shichang Joung
SK Telecom
Seoul, Republic of Korea
shichang.joung@sk.com

Bongmo Kim
SK Telecom
Seoul, Republic of Korea
bongmo.kim@sk.com

Yeul-Min Baek
SK Telecom
Seoul, Republic of Korea
ym.baek@sk.com

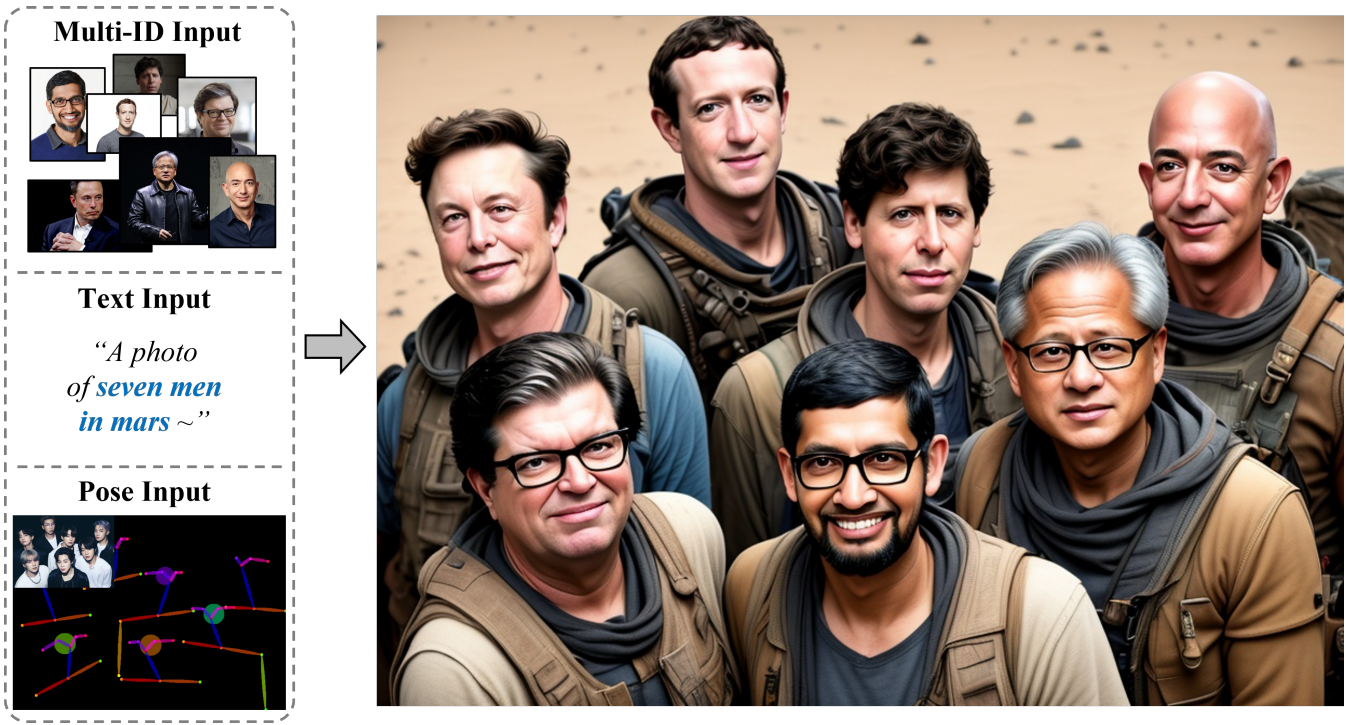

Figure 1: Result of Zero-shot Multi-ID Personalized Text-to-Image Generation: Our method generates an image (right) that satisfies all conditions, through a diffusion process when given images of multiple IDs, a text prompt, and an image for pose control(left).

## ABSTRACT

In the field of personalized image generation, the ability to create images preserving concepts has significantly improved. Creating an image that naturally integrates multiple concepts in a cohesive and visually appealing composition can indeed be challenging. This paper introduces "InstantFamily," an approach that employs a novel masked cross-attention mechanism and a multimodal embedding stack to achieve zero-shot multi-ID image generation. Our method effectively preserves ID as it utilizes global and local features from a pre-trained face recognition model integrated with text conditions. Additionally, our masked cross-attention mechanism enables the precise control of multi-ID and composition in the generated images. We demonstrate the effectiveness of InstantFamily through experiments showing its dominance in generating images with multi-ID, while resolving well-known multi-ID generation problems. Additionally, our model achieves state-of-the-art performance in both single-ID and multi-ID preservation. Furthermore, our model exhibits remarkable scalability with a greater number of ID preservation than it was originally trained with.

## 1 INTRODUCTION

The development of personalized image generation technologies has become feasible [7, 14, 25]. With the advent of such technologies,

the importance of accurately representing multi-ID within a single image has increased. These models often encounter limitations when attempting to effectively preserve the attributes of multiple IDs, thereby compromising on the individual characteristic that is essential for personalized imaging.

It is challenging to seamlessly integrate multiple identities to generate images. Problems such as identity mixing, where attributes from different individuals are incorrectly combined, complicate efforts to enhance the quality of personalized images. [39]. To address these challenges, this paper introduces "InstantFamily," a novel methodology using a masked cross-attention mechanism within the framework of a latent diffusion model. This approach not only preserves the identities of multiple individuals but also allows for the dynamic control of their poses and spatial relations. InstantFamily uses both global and local features in a face recognition model and employs masked cross-attention to effectively control identity. This enables the model to generate high performance in both identity accuracy and visual quality.

Our method, as shown in Fig. 1, can extend its applicability across various fields such as digital media, social platforms, and personalized content creation, thereby broadening the horizons of personalized image generation technology.

Our contributions are summarized as follows:

1) We propose a novel architecture that enables zero-shot multi-ID personalized text-to-image generation. Unlike other models that are limited to generating images of a fixed number of individuals, our model is scalable and capable of generating images featuring multiple persons.

2) Our approach achieves state-of-the-art performance in identity preservation. To ensure a fair comparison, we employed the same test data and evaluation code as FastComposer [39], which was previously regarded as the leading model in multi-ID personalized text-to-image generation.

3) A novel metric is proposed to evaluate identity preservation in scenarios involving multiple identities. Conventional metrics simply reiterate measurements for single identity preservation, failing to address the intricate issue of identity mixing that arises in multi-identity image generation. The introduced metric tackles this challenging problem, providing a comprehensive assessment of identity preservation when multiple identities are involved.

## 2 RELATED WORK

### 2.1 Text-to-Image Diffusion Models

The development of diffusion-based text-to-image generation models has rapidly progressed in recent years [8, 10, 19, 22, 23, 26, 31]. In particular, diffusion models that predict noise in the latent space have spurred research in this field and a variety of use cases [23]. However, generating images that meet the specific requirements of users solely based on text has its limitations, leading to research introducing various additional conditions to complement this.

ControlNet [41] represents a notable method for providing detailed guidance, such as locality and composition, in image generation tasks. It enables the incorporation of new conditions into the diffusion model through various preprocessed images, including sketches, depth maps, and human poses, offering localized information. Moreover, ControlNet is not dependent on any specific Stable Diffusion (SD) [23] model, making it generalizable to various SD implementations.

### 2.2 Personalization in Generation

Early personalized text-to-image generation methods involved training specific objects or styles by assigning concepts to words in prompts, like texture inversion [7] which optimizes embeddings. However, these methods are limited to the model's existing knowledge. DreamBooth [25] can fundamentally change the model for better personalization results but requires extensive training time and modifying the base model (UNet), potentially causing distribution shift when generating images from text. LoRA [11] adapters mitigate extensive training while preserving versatility. Recent studies present methods reducing training data needs or enabling personalized generation without training [12, 29, 37].

The IP-Adapter [40] is a method for prompting images to focus on a single subject. It can guide details that could not be guided by just text prompt through reference images. However, the IP-Adapter alone was insufficient for injecting image information into text comprehensively [34].

### 2.3 Single-ID Personalization

Similar to many image-to-image generation methods, an identity-preserving generation approach uses face image embedding in conjunction with text embedding to provide conditions for identity. For instance, PhotoMaker [17] employs a stacked ID embedding that merges image embedding extracted from a single person with class in text embedding that carries the meaning of ID. And PortraitBooth [20] augments the subject's features using TFace to imporve identity representation. However, due to its limited length, there is insufficient dimensionality to capture the full complexity of facial features. To improve the controllability and quality of identity-preserving generation, InstantID [34] utilizes ControlNet [41] along with face recognition feature.

To enhance image generation for preserving identity, methods utilizing features that excellently represent the face are explored [5, 33, 38]. Face0 [33] uses Inception-ResNet-V1 [32] trained on vggface2 [1], while DreamIdentity [4] employs a ViT-style [6] encoder. Particularly, Infinite-ID [38] captures identity information through the CLIP image encoder and mitigates interference from text prompts by deactivating cross-attention modules for text prompt. This method enhances both identity and semantic consistency effectively, yet lacks understanding of multi-ID and may generate artifacts when the human face occupies only a small part of the image.

### 2.4 Multi-ID Personalization

Models for single-ID personalization can be adapted to generate multi-ID images during inference by segmenting areas with masks. However, due to structural limitations, they cannot avoid problems related to multi-concept personalization, such as omitting, mixing, and splitting [39]. Due to these limitations, a method specifically designed for multi-ID personalization is required.

There are image generation methods targeting multi-concepts [9, 13–15, 18]. However, these methods mainly focus on fusion of weights, which requires either multiple conepts with clear separation [15] or fine-tuning of each concept [9, 13, 18]. Consequently, tuning-free text-to-image generation methods using multiple IDs are proposed [35, 39]. For example, FastComposer, similar to PhotoMaker [17], employs localizing cross-attention maps that link embedding extracted via an image encoder to text embedding. Despite using semantic maps to guide multi-ID, this method faces issues of identity loss due to compressed information and artifacts resulting from the blending of multiple individuals.

We propose InstantFamily which can preserve the identity of four or more individuals while freely controlling each pose. Our model achieves zero-shot generation of multi-ID images through the use of an embedding stack and a masked cross-attention mechanism.

## 3 PROPOSED METHOD

### 3.1 Preliminaries

The Latent Diffusion Model (LDM) [23] is a model that conducts the diffusion process in latent space and inputs text conditions through cross-attention [23]. In LDM, operations are performed in the latent space, leveraging each latent representation through a time-conditional UNet. Therefore, for each timestep with added noise, the corresponding latent representation is denoted as $z_t$. The autoencoder, represented by $\epsilon_\theta(z_t, t)$, is trained to predict the denoised variant of $z_t$.

$$\mathcal{L}_{\mathcal{LDM}} := \mathbb{E}_{\varepsilon(x), y, \epsilon \sim \mathcal{N}(0,1), t} \left[ \| \epsilon - \epsilon_\theta \left( z_t, t, \tau_\theta(y) \right) \|_2^2 \right], \quad (1)$$

In Eq. (1), $y$ is utilized as a condition to control the synthesis process, making use of texts, semantic maps, and similar resources. $\tau_\theta$ acts as a domain-specific encoder that takes our predefined condition $y$ as input and implements a cross-attention layer within the intermediate layers of the UNet. This approach diversifies the input modality. Equation 2 presents the formula for ControlNet [41], an end-to-end model designed to learn these task-specific conditions. Notably, while Eq. (1) characterize $\tau_\theta$ with a single input, in the context of ControlNet, it is described as accepting two distinct inputs as the text condition $c_t$ and the task-specific condition $c_f$.

$$\mathcal{L} = \mathbb{E}_{z_0, t, c_t, c_f, \epsilon} \left[ \left\| \epsilon - \epsilon_\theta \left( z_t, t, c_t, c_f \right) \right\|_2^2 \right], \quad (2)$$

Our method utilizes a face encoder to address these task-specific conditions, introducing a structure capable of stacking multiple IDs rather than a single ID.

### 3.2 InstantFamily

InstantFamily aims to generate splendid images while preserving identity, controlled on multi-person and multi-pose condition. Figure 2 shows our comprehensive architecture. The input to the proposed method integrates both global and local features from faces using the flattened one-dimensional face embeddings from the face encoder and two-dimensional last layer tensors from the same encoder. This dual input strategy ensures a rich capture of facial features, detailed further in Sec. 3.3

To effectively control multiple faces and bodies simultaneously, we designed masked cross-attention, which emphasizes face embedding while maintaining the integrity of text embedding. This approach involves concatenating multi-ID embedding with text embedding, where the mask contains information about the facial positions for each ID within the image. This mask assists in the weighted learning of positions corresponding to Text and IDs during the cross-attention process in UNet and ControlNet, elaborated in Sec. 3.4

The overall learning objective of the entire diffusion model is

$$\mathcal{L} = \mathbb{E}_{z_0, t, c_t, c_f, M, c_c, \epsilon} \left[ \left\| \epsilon - \epsilon_\theta \left( z_t, t, c_t, c_f, M, c_c \right) \right\|_2^2 \right], \quad (3)$$

where $c_f$ is the facial features, $M$ is the mask for facial feature and $c_c$ is the image condition for composition and poses of faces. Consequently, our method introduces a novel approach that allows for the preservation of multiple identities, while also providing the flexibility to freely control poses.

### 3.3 Multimodal Embedding Stack

To embed multiple IDs and integrate these multi-ID embedding $K_f$ with text embedding $K_t$, we introduce the embedding stack $K$. Figure 2(a) shows the process of detecting and embedding $N$ faces from the input image and then combining them with $K_t$. The overall formula for embedding stack $K$ is as follows.

$$K = ConCat(K_t, K_f) \quad (4)$$

$ConCat$ is a function of concatenation. Particularly, to embed multiple IDs from single image, we use face recognition encoder. We use two features for preserving identities, global feature $c_{gf} \in \mathbb{R}^{1 \times d_{gf}}$ and local feature $c_{lf} \in \mathbb{R}^{(L \times L) \times d_{lf}}$, to capture both abstract and detailed part of the face, where $d_{gf}$ and $d_{lf}$ are dimensions of global feature and local feature. $c_{gf}$ is a 1D face embedding feature that captures the global features of the face, and $c_{lf}$ is a 2D last layer's tensor of $n^{th}$ face and includes detailed local features of the face. Two facial features are combined through each projection process, which is expressed as $K_{fn} \in \mathbb{R}^{(L^2+1) \times d_K}$. This process is repeated for $N$ faces, and ultimately all $K_{fn}$ are concatenated.

$$K_f = ConCat(K_{f1}, K_{f2}, ..., K_{fN}) \quad (5)$$

The final $K$, which is the input of cross-attention, is created by stacking text embedding and multi-ID embedding, as shown in Eq. (4). Our embedding stack enables the application of text conditions while preserving multiple identities.

### 3.4 Masked Cross-Attention

To weight multi-ID embedding in all cross-attention layers, we introduce a masked cross-attention mechanism. Figure 2 (b) shows the process of our masked cross-attention mechanism. The overall formula for cross-attention operation is as follows.

$$Attention(Q, K, V) = softmax(\frac{M \odot QK^\top}{\sqrt{d_K}})V, \quad (6)$$

The masked cross-attention process unfolds in three steps, like Fig. 3. In the first step, the operation $QK^T$ is executed, which involves calculating the matrix multiplication between the query matrix $Q$ and the transposed key matrix $K^T$. In this operation, $Q$ is the matrix that rearranges the noisy image. The resulting matrix from this matrix multiplication encapsulates the compatibility scores between the elements of $Q$ and $K$, serving as a foundational component for the subsequent attention mechanism.

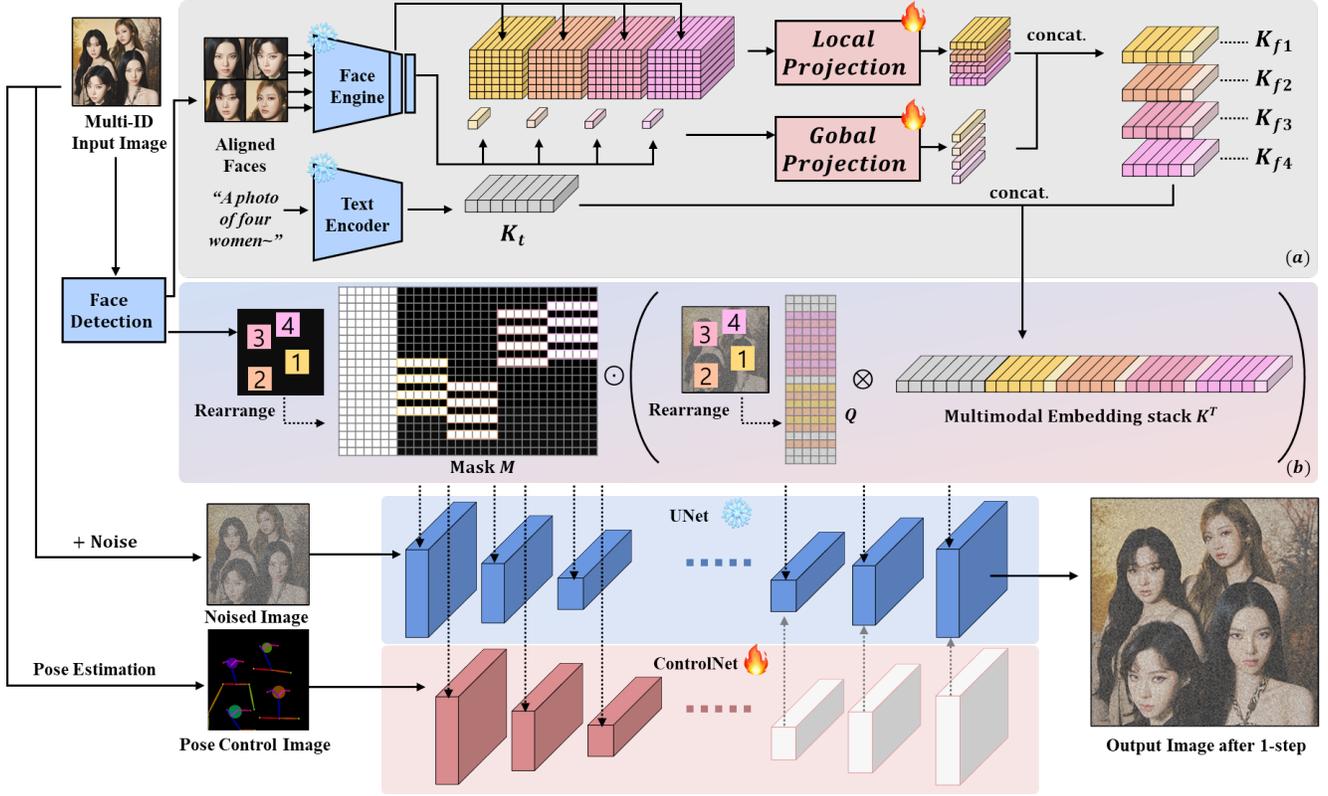

Figure 2: The proposed architecture of InstantFamily, (a) is the process of preparing a multimodal embedding stack, and (b) is the masked cross-attention mechanism. In (a), the multimodal embedding stack $K$ is obtained by integrating the face embedding $K_f$ and text embedding $K_t$. And the masked cross-attention process (b) is performed on UNet [24] and ControlNet [41]. Our mask $M$ has the effect of weighting id and text consistency.

In the second step, element-wise multiplication of $QK^T$ and mask $M$ is performed. Here, $\odot$ means this multiplication. At this time, Ours mask $M$ is defined as follows.

$$M = ConCat(M_t, M_{f1}, ..., M_{fN}) \qquad (7)$$

$M_t$ is a mask for the text prompt, and $M_f$ is a mask for N faces. This masking operation is essential for modulating the attention mechanism, which allows it to focus only on important parts of the input image. In other words, the mask $M$ allows us to control the attention process.

Finally, $Attention(Q, K, V)$ is calculated in the third step. This masked cross-attention is a process included in both UNet and ControlNet, enabling stable preservation of multi-ID.

### 3.5 Training and Inference
During training, faces that preserve identity are extracted from the same image, along with the mask and pose. As shown in Fig. 4, during inference, these conditions are not extracted from the same image. This means that different images can be used for the condition image of identity and the condition image for pose control. For pose control, the condition image can either be a real photo used as a reference or a generated image. Generally, in designing a service for content creation, the images for pose control could be predetermined along with text prompts, similar to pre-made templates. Then, by only inputting the images for identity, a structure can be created that generates multi-ID personalized images.

## 4 EXPERIMENTS
### 4.1 Dataset
We carefully selected approximately $2M$ images from the LAION-Face [28, 42] dataset, considering facial size and pose diversity. To effectively train on multi-ID, we collected $0.3M$ images featuring multiple people from the web and annotated them using BLIP2 [16] to add to our dataset. To minimize inter-class variation and facilitate the training of unique ID features, a portion of the facial recognition dataset was also included, which was similarly annotated using BLIP2. Additionally, to ensure high-quality image generation that accurately reflects prompts, we incorporated $50K$ generated images into the dataset.

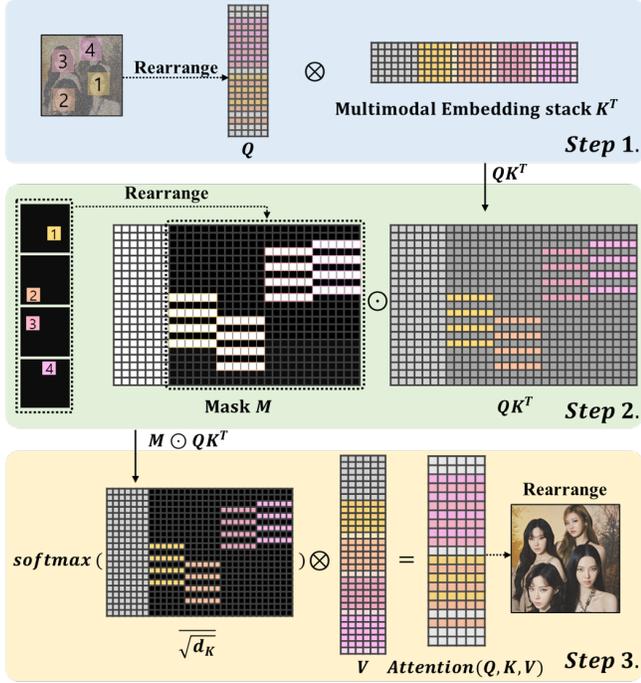

Figure 3: Masked Cross-Attention Process. Our cross-attention process is divided into three stages by applying a mask. *Step 1* is the process of obtaining $QK^T$ from the multimodal embedding stack $K^T$ and $Q$, *Step 2* is the process of applying the mask, and finally *Step 3* is the process of applying $Attention(Q, K, V)$.

### 4.2 Implementation Details

During training, the mask $m_{fn}$ within the 512 × 512 pixel-image space is defined as follows:

$$m_{fn} = \begin{cases} 1, & \text{if } n^{th} \text{ face area,} \\ 0, & \text{otherwise.} \end{cases} \qquad (8)$$

$M_{fn}$ is created by resizing $m_{fn}$ into 64×64, 32×32, 16×16 and 8×8, and then reshaping it at each stage of UNet for Stable Diffusion latent space. To reflect the overall features of the face and achieve harmony with the background and style, the face area utilized as a mask has been expanded by adding a 25% margin to the results of the face detector. OpenPose [2, 3, 30, 36] data was used for control, as it allows for control of both facial pose and body pose without facial shape restrictions, which makes it more suitable than the 5-point facial landmarks which is used from InstantID [34]. And we added color circles to the center of each face in the control image to guide the diffusion model to learn the attention location of each face feature to the corresponding face location. In learning how and where to reflect the ID in terms of order and position, the circle provides initial assistance, but ultimately, the influence of the mask is significant.

For training, $N$ is fixed to 4. For each images, $min(N, N')$ faces were randomly selected and their Face features were incorporated in a random order, where $N'$ denotes the total number of faces in the image. If fewer than four faces($N' < 4$) are inputted for training, bias may arise as only the embedding of the first facial position might be primarily used. Therefore, to ensure that face IDs are uniformly learned across the embedding order, the input facial IDs are stacked in a random order. In the case of $N' > 4$, pose information of all faces is incorporated, including those not used in training, to prevent unintended face generation in the background during inference. This means that even if trained on the poses of seven people, a maximum of four facial positions are circled. Faces without circles, meaning their embeddings are not stacked, are still used for pose control. This is because if this aspect is not learned, there is a higher likelihood of people being present in positions without poses, making it difficult to control the number of people. Figure 5 is an example of an image used in multi-face learning and its control image.

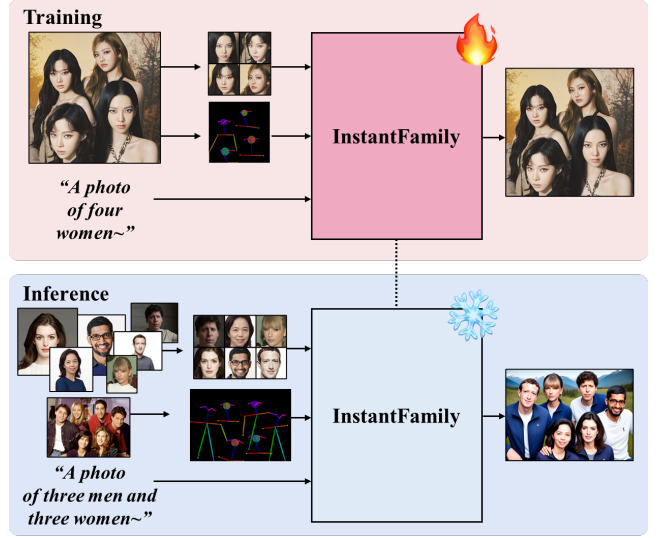

Figure 4: Training and Inference Process of InstantFamily. Once the model is trained, new inputs can be used for inference without retraining. Training requires multiple identity images paired with text prompts. However, during inference, users can use any multi-ID, and poses and text prompt can also be used flexibly.

### 4.3 Training

The base model which we used is SD1.5 and ControlNet is initailized with same parameteres. Training was conducted at a resolution of 512x512 using 8 NVIDIA A100(80GB) GPUs with a batch size of 8 per GPU. The total number of training steps was 400k. The rest of the hyperparameters were used as defaults from ControlNet [41].

### 4.4 Evaluation

For quantitative comparison, performance is typically evaluated using two metrics [38] in the field of personalized text-to-image Generation. The first is text consistency, which measures how well the generated images reflect the text prompts. This metric helps

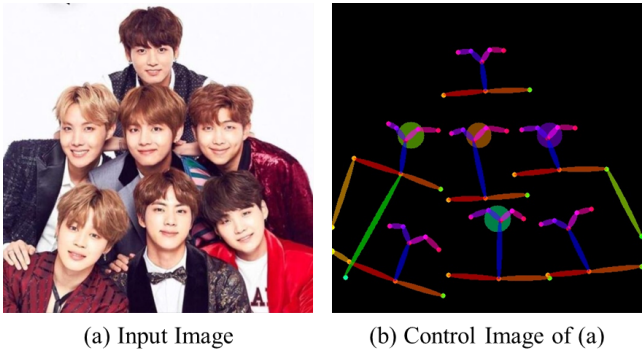

(a) Input Image      (b) Control Image of (a)

**Figure 5: Training image sample for multi-ID.** $N$ faces to be included in training are randomly selected at each step. Colored circles indicate randomly selected face positions.

**Table 1: Comparison of our method with state-of-the-art methods in generating images of single-ID. The best result is shown in underlined.**

| Method | Identity Preservation ↑ | Text Consistency ↑ |
| --- | --- | --- |
| IP-Adapter* [40] | 0.379 ± 0.246 | 0.250 ± 0.036 |
| InstantID [34] | 0.688 ± 0.249 | 0.228 ± 0.042 |
| PhotoMaker [17] | 0.445 ± 0.185 | 0.242 ± 0.040 |
| FastComposer [39] | 0.514 ± 0.189 | 0.229 ± 0.035 |
| InstantFamily(Ours) | 0.799 ± 0.086 | 0.205 ± 0.042 |

*IP-Adapter-FaceID-Plus-v2

to ensure that the performance of a pre-trained text-to-image generation model does not degrade through personalization training, that catastrophic forgetting does not occur, and that there is no overfitting to the ID. In the formula, CLIP's text encoder $E_{ct}$ is used for the prompt, and CLIP's image encoder $E_{ci}$ is used for the generated image. In the formula below, $P_t$ is the text prompt, $A$ is the ID Input image, and $A'$ is the image generated from $A$.

$$TextSim(P_t, A') = cos(E_{ct}(P_t), E_{ci}(A')) \qquad (9)$$

When calculating an indicator of how accurately the identity of an ID input image is preserved, some papers use the CLIP image encoder [21], while others use a face encoder. In this paper, we use FaceNet [27] to measure how consistently identities are preserved in the resulting images.

$$SingleSim(A, A') = cos(E_f(A), E_f(A')) \qquad (10)$$

It is very difficult in practice to numerically evaluate the generated images. Therefore, we computed it using the same ID input image, text prompt and code as FastComposer [39], which aims to generate multi-subject images. Even though the details of the generation process are private, we obtained values within the margin of error. According to the Tab. 1, Our method is much better at preserving identity. Our results on the text prompt are not the best, but they are not underperforming enough to be considered overfitting.

**Table 2: Comparison of our method with state-of-the-art methods in generating images of multi-ID. The best result is shown in underlined.**

| Method | Identity Preservation ↑ | Text Consistency ↑ |
| --- | --- | --- |
| FastComposer [39] | 1.392 ± 0.319 | 0.199 ± 0.039 |
| InstantFamily(Ours) | 1.620 ± 0.153 | 0.205 ± 0.050 |

$$MultiSim(A, B) = (SingleSim(A, A') + SingleSim(B, B'))/2 \\ + (1 - SingleSim(A', B')) \qquad (11)$$

While it is possible to evaluate multi-ID preservation ability simply by single-ID preservation capacity, it was not enough to assess challenging issues such as identity mixing inherent in multi-ID scenarios. Therefore, we propose a new metric capable of calculating multi-ID Similarity to assess this. The first term in the Eq. (11), averages the similarity between an image $A'$ generated from an ID input image $A$ and an image $B'$ generated from another ID input image $B$. This result refers to how consistently each single identity is preserved. Additionally, we address the mixing phenomenon by subtracting the similarity between the generated images $A'$ and $B'$ and incorporating the independence of IDs $A$ and $B$ into the multi-ID similarity calculation. For equal comparison, we chose FastComposer[39], which can generate multi-ID images at inference time without prior ID training or additional tools. As shown in Tab.2, our results achieve the highest performance. Especially, in the text consistency results of Tab.1, compared to comparison methods that have lower performance than Tab.2, our method maintains consistent results.

For qualitative comparison, as shown in Fig. 6, we compare our method on multi-ID image generation, FastComposer [39] and InstantID [34]. When using two people's ID images as input, IP-Adapter [40] applies the text condition, but the woman's jaw line is distorted. FastComposer applies the text condition, clothing style, to only one ID. When using ID images of 7 people as input, unlike existing methods, our method preserves the identity of all IDs.

### 4.5 Multi-ID Generation

In the provided Fig. 7, the multi-ID images generated by gradually increasing the number of people. It demonstrates that proposed single model can flexibly generate images for a variety of styles in a zero-shot manner, according to the number of people. To generate multi-ID images, we did not create additional "magic prompts" related to the photo composition but simply changed the prompt related to the number of people. For example, we can change a prompt from "1 man, a photo of a man~" to "4 men, a photo of four men~" to generate images with multiple people. Of course, to generate images with the desired IDs, it is necessary to stack the embeddings obtained through the face encoder and input the information related to the IDs.

Adjusting the number of IDs in multi-ID scenarios is a highly challenging task. Even when leveraging existing zero-shot single-ID personalized models and applying attention in a tricky manner to generate images, the conditioning can overlap to such an extent that the mixing issue becomes more pronounced, sometimes to the degree that it's impossible to generate normal images. Our method, having been trained up to stacking four IDs, can even generate images well for numbers exceeding four like Fig. 8. This implies that our approach utilizes a scalable embedding stack structure.

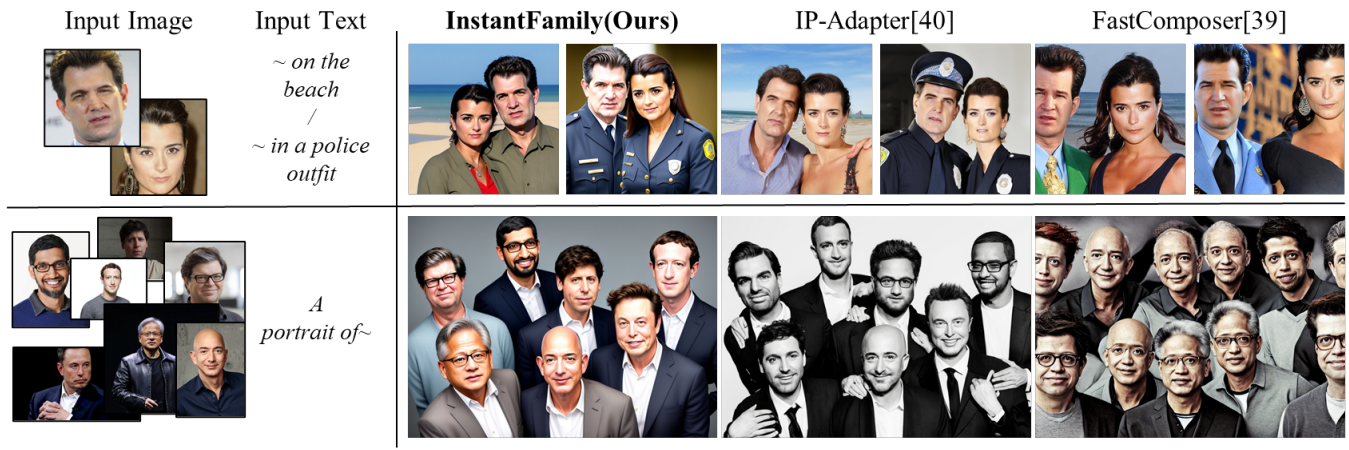

**Figure 6: Qualitative comparison of our method with state-of-the-art methods in generating images of multi-ID. The first row shows the results generated from two IDs and two versions of the text prompt. The second row shows the results generated from the seven IDs and a simple-style text prompt.**

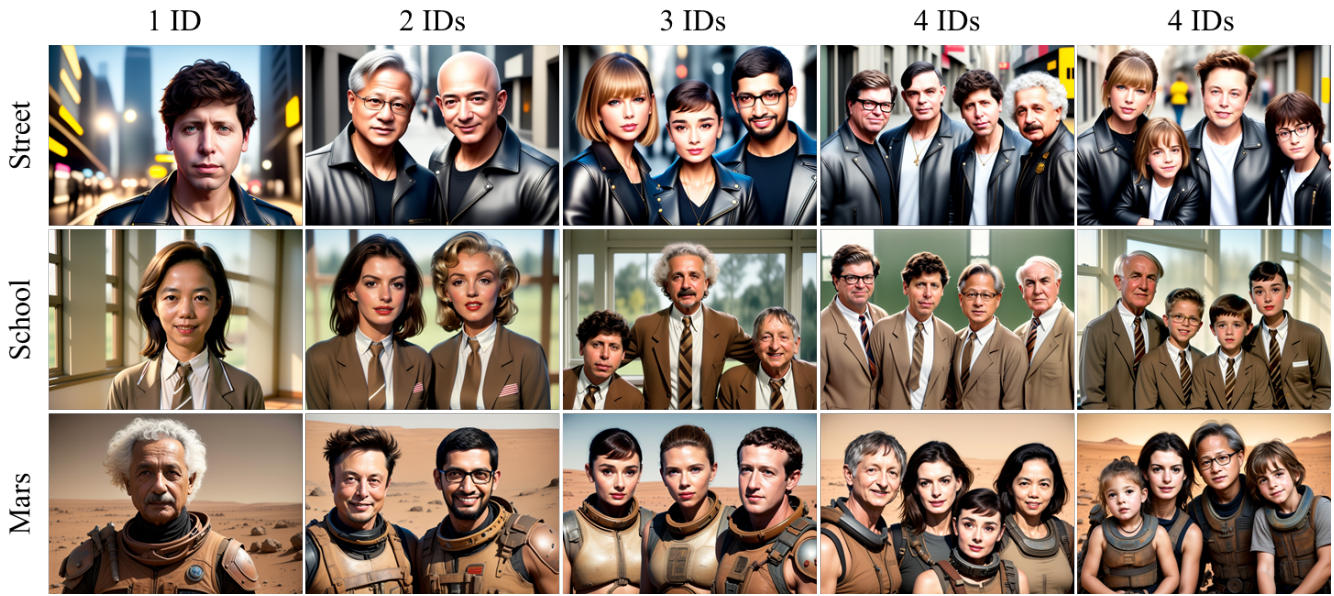

**Figure 7: Generation result from 1 ID to 4 IDs. Each row is the result generated for different numbers of input IDs and text prompt conditions such as "Mars", "School", and "Street".**

## 5 CONCLUSION

We proposed "InstantFamily," a significant advancement in the field of multi-ID personalized text-to-image generation. Furthermore, we introduced a new metric for quantitatively evaluating multi-ID preservation and demonstrated that our approach achieves state-of-the-art performance. Utilizing a scalable multimodal embedding stack and a masked cross-attention structure, we effectively trained on multi-ID scenarios. This enabled the generation of images that adhere to free composition and text prompts, thus enabling the production of new types of multimedia content that were previously unattainable.

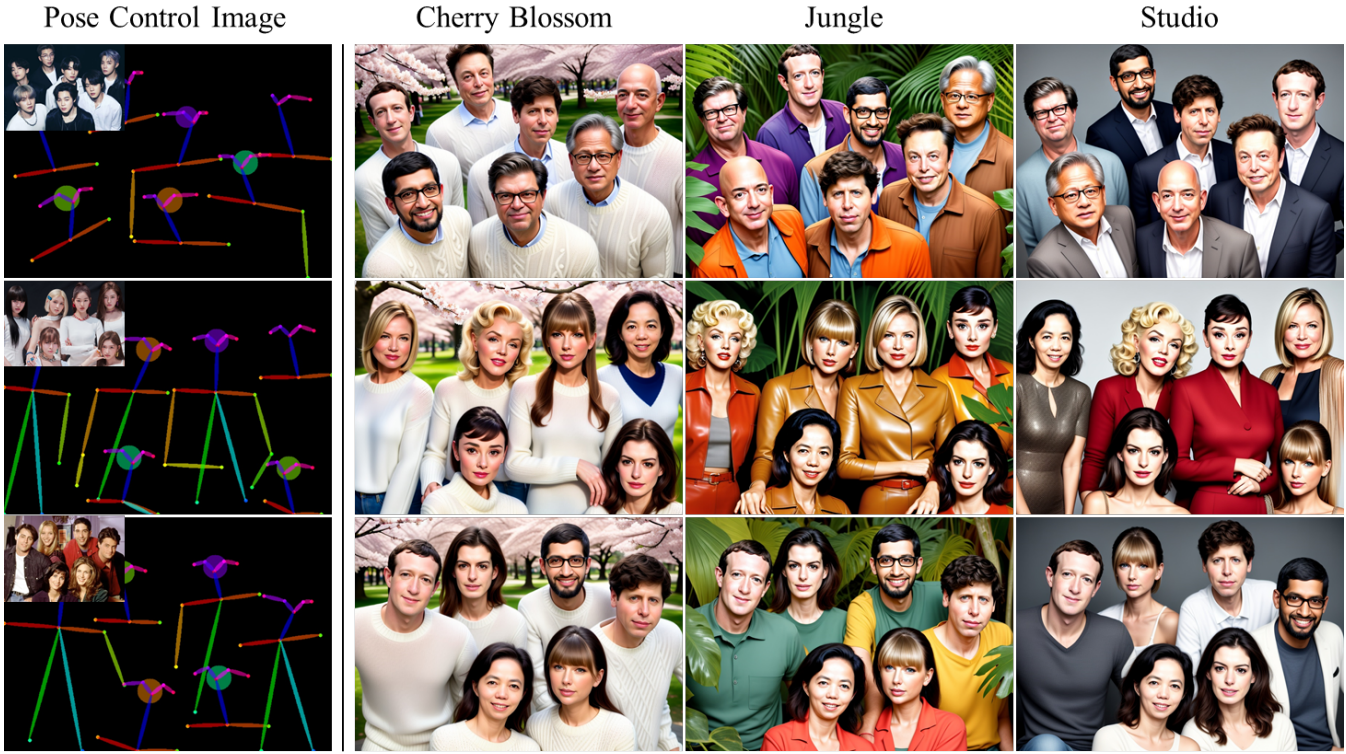

Figure 8: Generation result when $N > 4$. The first column is the pose control image used to generate each row result. The 2nd, 3rd, and 4th columns are the results generated with text prompt conditions such as "Cherry Blossom" and multi-ID images.

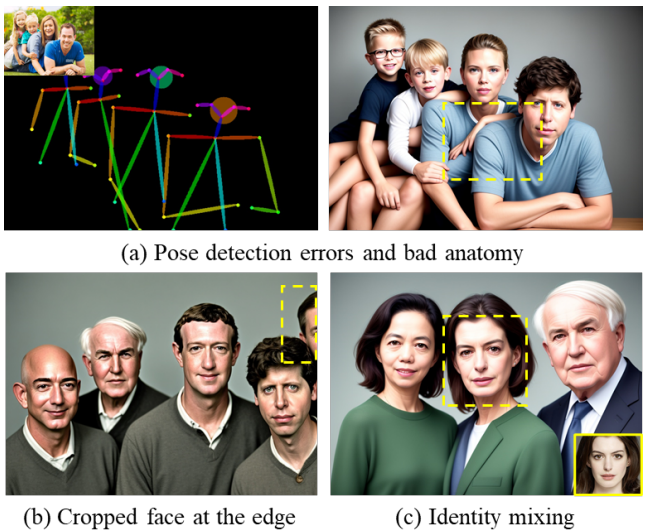

(a) Pose detection errors and bad anatomy

(b) Cropped face at the edge

(c) Identity mixing

Figure 9: Limitations. The yellow dashed lines in (a), (b), and (c) show the limitations of our method in some result images.

## 5.1 Limitations and Future Work

We have summarized our limitations in Fig. 9. The images used for pose control are predicted values through a pre-trained Open-Pose [2, 3, 30, 36] model. Therefore, it delivers performance that is dependent on this model, which can be a source of issues such as bad anatomy. We believe that a ControlNet-based model needs a structure that can mitigate this issue. Another challenge is the issue of faces appearing at the edges, as shown in Fig. 9. While we attempt to extract poses for all individuals present in the input images during training, it's challenging to do so with images similar to the examples generated, leading to such results. To be free from these issues, it is necessary to either refine the dataset or improve face detection and pose estimation. Lastly, there is the challenging problem of identity mixing, as depicted in Fig. 9. Although we believed improvements had been made, the problems still persisted. We assume that this is due to self-attention, but it is difficult to say that it is the only independent cause. There is a possibility to solve the problem by designing a structure that can independently calculate self-attention for each IDs. We will strive to find methodologies to solve the mentioned limitations and also hope to extend the proposed model to video generation.

## REFERENCES

[1] Qiong Cao, Li Shen, Weidi Xie, Omkar M Parkhi, and Andrew Zisserman. 2018. Vggface2: A dataset for recognising faces across pose and age. In *2018 13th IEEE international conference on automatic face & gesture recognition (FG 2018)*. IEEE,

67–74.
[2] Z. Cao, G. Hidalgo Martinez, T. Simon, S. Wei, and Y. A. Sheikh. 2019. OpenPose: Realtime Multi-Person 2D Pose Estimation using Part Affinity Fields. *IEEE Transactions on Pattern Analysis and Machine Intelligence* (2019).
[3] Zhe Cao, Tomas Simon, Shih-En Wei, and Yaser Sheikh. 2017. Realtime Multi-Person 2D Pose Estimation using Part Affinity Fields. In *CVPR*.
[4] Zhuowei Chen, Shancheng Fang, Wei Liu, Qian He, Mengqi Huang, and Zhendong Mao. 2024. DreamIdentity: Enhanced Editability for Efficient Face-Identity Preserved Image Generation. In *Proceedings of the AAAI Conference on Artificial Intelligence*, Vol. 38. 1281–1289.
[5] Zhuowei Chen, Shancheng Fang, Wei Liu, Qian He, Mengqi Huang, Yongdong Zhang, and Zhendong Mao. 2023. Dreamidentity: Improved editability for efficient face-identity preserved image generation. *arXiv preprint arXiv:2307.00300* (2023).
[6] Alexey Dosovitskiy, Lucas Beyer, Alexander Kolesnikov, Dirk Weissenborn, Xiaohua Zhai, Thomas Unterthiner, Mostafa Dehghani, Matthias Minderer, Georg Heigold, Sylvain Gelly, et al. 2020. An Image is Worth 16x16 Words: Transformers for Image Recognition at Scale. In *International Conference on Learning Representations*.
[7] Rinon Gal, Yuval Alaluf, Yuval Atzmon, Or Patashnik, Amit Haim Bermano, Gal Chechik, and Daniel Cohen-or. 2022. An Image is Worth One Word: Personalizing Text-to-Image Generation using Textual Inversion. In *The Eleventh International Conference on Learning Representations*.
[8] Rinon Gal, Moab Arar, Yuval Atzmon, Amit H Bermano, Gal Chechik, and Daniel Cohen-Or. 2023. Encoder-based domain tuning for fast personalization of text-to-image models. *ACM Transactions on Graphics (TOG)* 42, 4 (2023), 1–13.
[9] Yuchao Gu, Xintao Wang, Jay Zhangjie Wu, Yujun Shi, Yunpeng Chen, Zihan Fan, Wuyou Xiao, Rui Zhao, Shuning Chang, Weijia Wu, et al. 2024. Mix-of-show: Decentralized low-rank adaptation for multi-concept customization of diffusion models. *Advances in Neural Information Processing Systems* 36 (2024).
[10] Jonathan Ho, Ajay Jain, and Pieter Abbeel. 2020. Denoising diffusion probabilistic models. *Advances in neural information processing systems* 33 (2020), 6840–6851.
[11] Edward J Hu, Phillip Wallis, Zeyuan Allen-Zhu, Yuanzhi Li, Shean Wang, Lu Wang, Weizhu Chen, et al. 2021. LoRA: Low-Rank Adaptation of Large Language Models. In *International Conference on Learning Representations*.
[12] Yuval Kirstain, Omer Levy, and Adam Polyak. 2023. X&fuse: Fusing visual information in text-to-image generation. *arXiv preprint arXiv:2303.01000* (2023).
[13] Zhe Kong, Yong Zhang, Tianyu Yang, Tao Wang, Kaihao Zhang, Bizhu Wu, Guanying Chen, Wei Liu, and Wenhan Luo. 2024. OMG: Occlusion-friendly Personalized Multi-concept Generation in Diffusion Models. *arXiv preprint arXiv:2403.10983* (2024).
[14] Nupur Kumari, Bingliang Zhang, Richard Zhang, Eli Shechtman, and Jun-Yan Zhu. 2023. Multi-concept customization of text-to-image diffusion. In *Proceedings of the IEEE/CVF Conference on Computer Vision and Pattern Recognition*. 1931–1941.
[15] Nupur Kumari, Bingliang Zhang, Richard Zhang, Eli Shechtman, and Jun-Yan Zhu. 2023. Multi-Concept Customization of Text-to-Image Diffusion. (2023).
[16] Junnan Li, Dongxu Li, Silvio Savarese, and Steven Hoi. 2023. Blip-2: Bootstrapping language-image pre-training with frozen image encoders and large language models. In *International conference on machine learning*. PMLR, 19730–19742.
[17] Zhen Li, Mingdeng Cao, Xintao Wang, Zhongang Qi, Ming-Ming Cheng, and Ying Shan. 2024. PhotoMaker: Customizing Realistic Human Photos via Stacked ID Embedding. In *IEEE Conference on Computer Vision and Pattern Recognition (CVPR)*.
[18] Zhiheng Liu, Ruili Feng, Kai Zhu, Yifei Zhang, Kecheng Zheng, Yu Liu, Deli Zhao, Jingren Zhou, and Yang Cao. 2023. Cones: Concept Neurons in Diffusion Models for Customized Generation. In *International Conference on Machine Learning*. PMLR, 21548–21566.
[19] Alexander Quinn Nichol, Prafulla Dhariwal, Aditya Ramesh, Pranav Shyam, Pamela Mishkin, Bob Mcgrew, Ilya Sutskever, and Mark Chen. 2022. GLIDE: Towards Photorealistic Image Generation and Editing with Text-Guided Diffusion Models. In *International Conference on Machine Learning*. PMLR, 16784–16804.
[20] Xu Peng, Junwei Zhu, Boyuan Jiang, Ying Tai, Donghao Luo, Jiangning Zhang, Wei Lin, Taisong Jin, Chengjie Wang, and Rongrong Ji. 2023. PortraitBooth: A Versatile Portrait Model for Fast Identity-preserved Personalization. *arXiv preprint arXiv:2312.06354* (2023).
[21] Alec Radford, Jong Wook Kim, Chris Hallacy, Aditya Ramesh, Gabriel Goh, Sandhini Agarwal, Girish Sastry, Amanda Askell, Pamela Mishkin, Jack Clark, et al. 2021. Learning transferable visual models from natural language supervision. In *International conference on machine learning*. PMLR, 8748–8763.
[22] Aditya Ramesh, Mikhail Pavlov, Gabriel Goh, Scott Gray, Chelsea Voss, Alec Radford, Mark Chen, and Ilya Sutskever. 2021. Zero-shot text-to-image generation. In *International conference on machine learning*. Pmlr, 8821–8831.
[23] Robin Rombach, Andreas Blattmann, Dominik Lorenz, Patrick Esser, and Björn Ommer. 2022. High-resolution image synthesis with latent diffusion models. In *Proceedings of the IEEE/CVF conference on computer vision and pattern recognition*. 10684–10695.
[24] Olaf Ronneberger, Philipp Fischer, and Thomas Brox. 2015. U-net: Convolutional networks for biomedical image segmentation. In *Medical image computing and computer-assisted intervention–MICCAI 2015: 18th international conference, Munich, Germany, October 5-9, 2015, proceedings, part III 18*. Springer, 234–241.
[25] Nataniel Ruiz, Yuanzhen Li, Varun Jampani, Yael Pritch, Michael Rubinstein, and Kfir Aberman. 2023. Dreambooth: Fine tuning text-to-image diffusion models for subject-driven generation. In *Proceedings of the IEEE/CVF Conference on Computer Vision and Pattern Recognition*. 22500–22510.
[26] Chitwan Saharia, William Chan, Saurabh Saxena, Lala Li, Jay Whang, Emily L Denton, Kamyar Ghasemipour, Raphael Gontijo Lopes, Burcu Karagol Ayan, Tim Salimans, et al. 2022. Photorealistic text-to-image diffusion models with deep language understanding. *Advances in neural information processing systems* 35 (2022), 36479–36494.
[27] Florian Schroff, Dmitry Kalenichenko, and James Philbin. 2015. FaceNet: A unified embedding for face recognition and clustering. In *2015 IEEE Conference on Computer Vision and Pattern Recognition (CVPR)*. IEEE. https://doi.org/10.1109/cvpr.2015.7298682
[28] Christoph Schuhmann, Richard Vencu, Romain Beaumont, Robert Kaczmarczyk, Clayton Mullis, Aarush Katta, Theo Coombes, Jenia Jitsev, and Aran Komatsuzaki. 2021. Laion-400m: Open dataset of clip-filtered 400 million image-text pairs. *arXiv preprint arXiv:2111.02114* (2021).
[29] Jing Shi, Wei Xiong, Zhe Lin, and Hyun Joon Jung. 2023. Instantbooth: Personalized text-to-image generation without test-time finetuning. *arXiv preprint arXiv:2304.03411* (2023).
[30] Tomas Simon, Hanbyul Joo, Iain Matthews, and Yaser Sheikh. 2017. Hand Keypoint Detection in Single Images using Multiview Bootstrapping. In *CVPR*.
[31] Yang Song, Jascha Sohl-Dickstein, Diederik P Kingma, Abhishek Kumar, Stefano Ermon, and Ben Poole. 2020. Score-Based Generative Modeling through Stochastic Differential Equations. In *International Conference on Learning Representations*.
[32] Christian Szegedy, Sergey Ioffe, Vincent Vanhoucke, and Alexander Alemi. 2017. Inception-v4, inception-resnet and the impact of residual connections on learning. In *Proceedings of the AAAI conference on artificial intelligence*, Vol. 31.
[33] Dani Valevski, Danny Lumen, Yossi Matias, and Yaniv Leviathan. 2023. Face0: Instantaneously conditioning a text-to-image model on a face. In *SIGGRAPH Asia 2023 Conference Papers*. 1–10.
[34] Qixun Wang, Xu Bai, Haofan Wang, Zekui Qin, and Anthony Chen. 2024. Instantid: Zero-shot identity-preserving generation in seconds. *arXiv preprint arXiv:2401.07519* (2024).
[35] Yibin Wang, Weizhong Zhang, Jianwei Zheng, and Cheng Jin. 2023. High-fidelity Person-centric Subject-to-Image Synthesis. *arXiv preprint arXiv:2311.10329* (2023).
[36] Shih-En Wei, Varun Ramakrishna, Takeo Kanade, and Yaser Sheikh. 2016. Convolutional pose machines. In *CVPR*.
[37] Yuxiang Wei, Yabo Zhang, Zhilong Ji, Jinfeng Bai, Lei Zhang, and Wangmeng Zuo. 2023. Elite: Encoding visual concepts into textual embeddings for customized text-to-image generation. In *Proceedings of the IEEE/CVF International Conference on Computer Vision*. 15943–15953.
[38] Yi Wu, Ziqiang Li, Heliang Zheng, Chaoyue Wang, and Bin Li. 2024. Infinite-ID: Identity-preserved Personalization via ID-semantics Decoupling Paradigm. *arXiv preprint arXiv:2403.11781* (2024).
[39] Guangxuan Xiao, Tianwei Yin, William T Freeman, Frédo Durand, and Song Han. 2023. Fastcomposer: Tuning-free multi-subject image generation with localized attention. *arXiv preprint arXiv:2305.10431* (2023).
[40] Hu Ye, Jun Zhang, Sibo Liu, Xiao Han, and Wei Yang. 2023. Ip-adapter: Text compatible image prompt adapter for text-to-image diffusion models. *arXiv preprint arXiv:2308.06721* (2023).
[41] Lvmin Zhang, Anyi Rao, and Maneesh Agrawala. 2023. Adding conditional control to text-to-image diffusion models. In *Proceedings of the IEEE/CVF International Conference on Computer Vision*. 3836–3847.
[42] Yinglin Zheng, Hao Yang, Ting Zhang, Jianmin Bao, Dongdong Chen, Yangyu Huang, Lu Yuan, Dong Chen, Ming Zeng, and Fang Wen. 2022. General facial representation learning in a visual-linguistic manner. In *Proceedings of the IEEE/CVF Conference on Computer Vision and Pattern Recognition*. 18697–18709.